\documentclass[final]{cvpr}

\usepackage{balance}
\usepackage{times}
\usepackage{epsfig}
\usepackage{graphicx}
\usepackage{amsmath}
\usepackage{amssymb}
\usepackage[labelfont=bf]{caption}

\usepackage[pagebackref=true,breaklinks=true,colorlinks,bookmarks=false]{hyperref}

\def\cvprPaperID{****} 
\def\confYear{CVPR 2021}

\begin{document}

\title{The Challenge of Appearance-Free Object Tracking \\ with Feedforward Neural Networks}




\author{Girik Malik\\
Northeastern U.\\
{\tt\small gmalik@ccs.neu.edu}

\and
Drew Linsley \hspace{6mm} Thomas Serre\\
Brown University\\
{\tt\small drew\_linsley, thomas\_serre@brown.edu}

\and
Ennio Mingolla\\
Northeastern U.\\
{\tt\small e.mingolla@northeastern.edu}
}

\maketitle

\begin{abstract}
   Nearly all models for object tracking with artificial neural networks depend on appearance features extracted from a ``backbone'' architecture, designed for object recognition. Indeed, significant progress on object tracking has been spurred by introducing backbones that are better able to discriminate objects by their appearance. However, extensive neurophysiology and psychophysics evidence suggests that biological visual systems track objects using both appearance and motion features~\cite{Andersen1997-lo}. Here, we introduce \textit{Pathtracker}, a visual challenge inspired by cognitive psychology, which tests the ability of observers to learn to track objects solely by their motion. We find that standard 3D-convolutional deep network models struggle to solve this task when clutter is introduced into the generated scenes, or when objects travel long distances. This challenge reveals that tracing the path of object motion is a blind spot of feedforward neural networks. We expect that strategies for appearance-free object tracking from biological vision can inspire solutions these failures of deep neural networks.
   
\end{abstract}


\section{The \textit{Pathtracker} Challenge}
\label{sec:challenge}
\begin{figure}[t]
\begin{center}\small
\includegraphics[width=.4\textwidth]{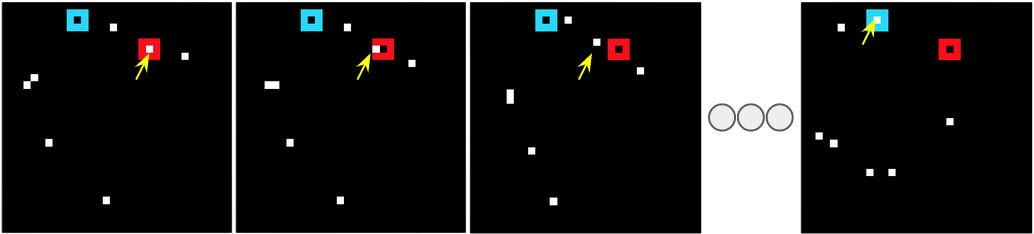}
\end{center}
\vspace{-4mm}
\caption{A positive example from the \textit{Pathtracker} challenge. The target dot starts from inside the red marker on the first frame, and ends inside the blue marker on the last frame. In this example, there are 6 distractor dots, which are identical to the target and follow random trajectories (see Fig.~\ref{fig:fig2}; note that the number of distractors is a free parameter that we vary in our benchmarks). The first three frames show the path of the target dot that began inside the red square, denoted by yellow arrows (these annotations are not visible in the challenge). All moving objects are identical, making it impossible to distinguish the target and distractors by appearance. Thus, observers must track the target from the first frame until the end of the sequence. }\vspace{-4mm}
\label{fig:fig1}
\end{figure}

We introduce a novel synthetic visual tracking challenge, called \textit{Pathtracker}. The challenge, inspired by the multi-object tracking paradigm of cognitive psychology \cite{pylyshynandstorm, blaser2000tracking}, is designed to test the ability of observers to learn to track an object in the presence of clutter over long distances (Fig.~\ref{fig:fig1}). \textit{Pathtracker} joins a growing body of work which uses visually simple challenges, inspired by cognitive science, to identify limitations of feedforward models and inspire solutions that push the state of the art in computer vision~\cite{NEURIPS2020_766d856e,Linsley2019-gn}.

\begin{figure}[t]
\begin{center}\small
\includegraphics[width=0.23\textwidth]{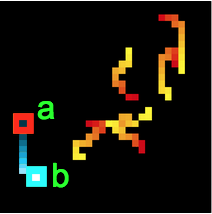}
\includegraphics[width=0.233\textwidth]{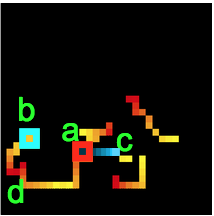}
\end{center}
\vspace{-4mm}
\caption{A diagram of dot motion for a positive versus a negative video in the \textit{Pathtracker} challenge. Motion is plotted by superimposing the frames in each video and coloring the trajectories of the targets and distractors. In the positive example on the left, the target dot starts inside red marker (\textit{a}) and ends inside the blue marker (\textit{b}). The target's path is colored in blue, whereas distractor paths are colored in yellow/red. In the negative example on the right, the target dot starts inside red marker (\textit{a}) but a distractor dot (\textit{d}) ends inside the blue marker (\textit{b}). The target dot instead ends at an arbitrary location (\textit{c}). The target often crosses paths with distractors, forcing observers to handle occlusion in individual frames and making it difficult for neural architectures to learn shortcut solutions to the challenge.}
\label{fig:fig2}
\end{figure}

The goal of \textit{Pathtracker} is to determine whether or not a ``target'' dot starting on a red marker travels to a blue ``finish'' marker. These two markers are stationary and placed at random positions in each video. The target and distractor dots are white squares that follow procedurally-generated trajectories (Fig.~\ref{fig:fig2}; see \S\ref{subsec:speed} for details). In positive examples, the target dot finishes its trajectory in the blue marker at the end of the video. In negative examples, a distractor dot, finishes its trajectory on the blue marker, while the target dot ends up somewhere else.

In the most basic version of \textit{Pathtracker}, observers have to track the target over a 32-frame video, while ignoring a single distractor. The task is made harder by increasing (\textit{i}) the number of distractors, (\textit{ii}) the length of the video, or (\textit{iii}) the speed of the target and distractors. Increasing the number of distractors makes it more likely that target and distractors cross -- requiring an observer to distinguish between identical objects and bind identity with motion trajectories. Gestalt principles such as the principle of continuity may provide robust cues to solve such challenge but it is unknown whether modern feedforward neural networks are capable of learning such spatiotemporal features. 


\section{Stimulus design}
\label{subsec:speed}
Challenge videos consist of 32, 64 or 128 frames of $32\times32$ pixels. Target and distractor trajectories are curved and variably shaped. On two successive frames, the displacement between the coordinate positions of the dot is constrained to be not more than 2 pixels and to not bend more than $20^\circ$. This means that all dots look like they are  meandering through the scene, never turning at acute angles.  Start (red) and finish (blue) markers are placed at the starting and ending positions of the target in positive examples. In negative examples, these markers are placed at the start of the target and the end of a randomly selected distractor. Targets and distractors move at the same speed in all videos. In our challenge we produce videos with one of three possible speeds: normal, fast, or very fast. Targets and distractors traverse longer trajectories as speed is increased, but for every speed the target always ends in the finish marker on positive examples, and a distractor ends in the finish marker on negative examples.

\section{Benchmark Models: 3D-CNNs}

Fiaz et al. \cite{fiaz2018} noted that deep learning-based tracking algorithms rely on appearance: they use the correlation between appearance-based feature maps of successive frames to track target objects. More recent methods have shown incremental improvements by using this same approach on object-proposals from instance segmentation models applied to individual frames~\cite{bertasius2020maskprop}. The number of deep network architectures that can be used for appearance-free video processing are fairly limited. Although tracking is a well studied subfield in computer vision, to our knowledge there are no deep learning-based methods for ``feature agnostic'' tracking based on motion.

We focus our experiments on Inception 3D (I3D) networks~\cite{carreira2017}, which are a standard approach to action recognition in natural videos. We split \textit{Pathtracker} dataset into separate train/test folds, and perform a large-scale search over I3D hyperparameters to optimize its performance. In total, we explore the following hyperparameters: (\textit{i}) random weight initializations vs. pretrained Kinetics initializations, (\textit{ii}) learning rates of \{$1\times10^{-4}$, $3\times10^{-4}$, $1\times10^{-5}$, $1\times10^{-6}$\}, and (\textit{iii}) $\mathcal{L}_2$ weight decay scaled at \{$4\times10^{-5}$, $1\times10^{-6}$\}. All models were trained for $1000$ epochs on batches of 64 videos. Here, we report best performance across our hyperparameter search. (See section \ref{sec:rgbhypersearch} in supplementary materials for details of our experiments with hyperparameter search.) 

Every variant of \textit{Pathtracker} had 20K training and 20K test samples. Models were trained on TPUs and evaluated on versions of the challenge with different numbers of distractors, frames, and speeds. We also include experiments where I3D was trained on optic flow encodings of the datasets (see supplementary material). 

\begin{table}
\centering
\caption{I3D evaluation accuracy on variable length \textit{Pathtracker} datasets of ``normal'' speed.}
\label{tab:table1}
\begin{tabular}{llll}
                              & \multicolumn{3}{l}{\textbf{Video Dimension (D,H,W)}}        \\
\textbf{\textbf{Distractors}} & \textbf{32,32,32} & \textbf{64,32,32} & \textbf{128,32,32}  \\
\textbf{1}                    & 93.88             & 81.29             & 58.20               \\
\textbf{6}                    & 86.90             & 67.39             & 56.40               \\
\textbf{15}                   & 83.59             & 65.13             & 56.55               \\
\textbf{26}                   & 81.03             & 64.29             & 56.06              
\end{tabular}
\end{table}

\begin{table}
\caption{I3D evaluation accuracy on \textit{Pathtracker} datasets with variable dot speed. Dots travel longer distances when speed is increased above ``normal''. We generated ``fast'' and ``very fast'' versions of \textit{Pathtracker} videos, in which dots traveled 2$\times$ or 4$\times$ their normal distance, respectively. We set the maximum path length to be consistent with dot path length in the 128-frame videos; hence, we do not generate ``very fast'' versions of the 64$\times$32$\times$32 datasets. See \S \ref{subsec:speed} for details on speed variations.}
\label{tab:table2}
\centering
\begin{tabular}{lllll}
                              & \multicolumn{4}{l}{\textbf{Video Dimension (D,H,W)}}                                                       \\
\textbf{\textbf{Dist.}} & \textbf{32,32,32} & \textbf{\textbf{32,32,32}} & \textbf{\textbf{64,32,32}} & \textbf{\textbf{128,32,32}}  \\
\textbf{Speed}                & \textbf{Fast}     & \textbf{Very Fast}         & \textbf{Fast}              & \textbf{Normal}              \\
\textbf{1}                    & 81.50             & 59.82                      & 59.79                      & 56.66                        \\
\textbf{6}                    & 68.13             & 55.56                      & 57.30                      & 56.40                        \\
\textbf{15}                   & 63.04             & 55.40                      & 56.12                      & 56.54                        \\
\textbf{26}                   & 62.89             & 51.10                      & 51.62                      & 50.81                       
\end{tabular}
\end{table}

\section{Results}
I3D performed well on a 1-distractor, 32-frame version of mixed-channel version of \textit{Pathtracker} dataset (Fig.~\ref{fig:fig1}), reaching 93\% accuracy. However, I3D's performance monotonically decreased as clutter and tracking length increased (Table \ref{tab:table1}). Indeed, we found a sharp drop in I3D's performance for 64-frame videos when more than one distractor was included (see the drop from 1$\rightarrow{6}$ in Table \ref{tab:table1}). I3D also struggled to learn the task, even when only a single distractor was present, in longer 128-frame videos.

We further found that I3D struggled to learn \textit{Pathtracker} when we increased the speed of dots in the videos. I3D was unable to break $60\%$ accuracy when dots moved ``very fast'' in a 32-frame video or ``fast'' in a 64-frame video (see Table \ref{tab:table2}).

One possible explanation why I3D struggled in so many conditions of the \textit{Pathtracker} challenge is that it was either over- or under-parameterized, which caused unstable learning. To address this, we repeated our experiments with wider and narrower I3D networks, containing 4$\times$ and $\frac{1}{4}th$ the number of features as the original model. 

\noindent\textbf{Narrower Network: } We reduced the feature maps on all but readout layers to $1/4^{th}$ of their original size. We observed a monotonic decrease in accuracy, as in case of original feature maps. We observe a 4-5\% increase in accuracy on the datasets with path lengths corresponding to a 128-frame task and 26 distractors. Accuracy on other datasets remain constant. See tables \ref{tab:narrow1} and \ref{tab:narrow2} for constant and variable speed experiments respectively. 

\begin{table}[ht]
\caption{Evaluation accuracy of I3D with narrower network with variation in the number of frames.}
\vspace{-4mm}
\label{tab:narrow1}
\begin{center}\small
\begin{tabular}{llll}
&  \multicolumn{3}{l}{\textbf{Video dimension (D,H,W)}} \\
\textbf{Distractors} & \textbf{32,32,32}                & \textbf{64,32,32}                       & \textbf{128,32,32}                       \\
\textbf{1 dist}         & 92.53              & 80.44                      & 56.71                      \\ 
\textbf{6 dist}         & 84.58              & 64.45                     & 56.80                      \\ 
\textbf{15 dist}        & 80.11              & 64.19                     & 56.18                      \\ 
\textbf{26 dist}        & 79.10              & 63.28 & 55.70 \\ 
\end{tabular}
\end{center}
\vspace{-4mm}
\end{table}

\begin{table}[ht]
\caption{Evaluation accuracy of I3D with narrower network with variable speed datasets, compared to the longer constant speed version.}
\vspace{-4mm}
\label{tab:narrow2}
\begin{center}\small
\begin{tabular}{lllll}
&  \multicolumn{4}{l}{\textbf{Video dimension (D,H,W)}} \\
\textbf{Dist.} & \textbf{32,32,32} & \textbf{32,32,32} & \textbf{64,32,32} & \textbf{128,32,32} \\
\textbf{Speed} & \textbf{Fast} & \textbf{Very Fast} & \textbf{Fast} & \textbf{Normal}              \\
\textbf{1 dist}         & 79.16 & 57.64 & 58.08 & 56.71 \\ 
\textbf{6 dist}         & 66.61 & 55.79 & 57.21 & 56.80 \\ 
\textbf{15 dist}        & 64.35 & 55.69 & 56.08 & 56.18 \\ 
\textbf{26 dist}        & 62.59 & 55.35 & 56.25 & 55.70 \\ 
\end{tabular}
\end{center}
\vspace{-4mm}
\end{table}

\noindent\textbf{Wider Network: } We increased the feature maps on all but readout layers to $4\times$ of their original size. We again observed a similar monotonic decrease in accuracy, as in case of experiments with the original feature maps. Contrary to the belief of large feature maps being able to learn better representation of objects, we observe a 14\% drop in accuracy for 64-frame 26 distractor version of the dataset. For other longer path length datasets, we also observed small drops in accuracy. Only the shorter path length videos show an insignificant increase in accuracy. See tables \ref{tab:wide1} and \ref{tab:wide2} for constant and variable speed experiments respectively. 

\begin{table}[ht]
\caption{Evaluation accuracy of I3D with wider network with variation in the number of frames.}
\vspace{-4mm}
\label{tab:wide1}
\begin{center}\small
\begin{tabular}{llll}
&  \multicolumn{3}{l}{\textbf{Video dimension (D,H,W)}} \\
\textbf{Distractors}               & \textbf{32,32,32}                & \textbf{64,32,32}                       & \textbf{128,32,32}                       \\
\textbf{1 dist}         & 93.42              & 79.56                      & 53.59                      \\ 
\textbf{6 dist}         & 87.28              & 64.40                     & 52.68                      \\ 
\textbf{15 dist}        & 85.23              & 64.49                     & 50.77                      \\ 
\textbf{26 dist}        & 82.86              & 50.95 & 50.75 \\ 
\end{tabular}
\end{center}
\vspace{-4mm}
\end{table}

\begin{table}[ht]
\caption{Evaluation accuracy of I3D with wider network with variable speed datasets, compared to the longer constant speed version.}
\vspace{-4mm}
\label{tab:wide2}
\begin{center}\small
\begin{tabular}{lllll}
&  \multicolumn{4}{l}{\textbf{Video dimension (D,H,W)}} \\
\textbf{Dist.} & \textbf{32,32,32} & \textbf{32,32,32} & \textbf{64,32,32} & \textbf{128,32,32} \\
\textbf{Speed}                & \textbf{Fast}     & \textbf{Very Fast}         & \textbf{Fast}              & \textbf{Normal}              \\
\textbf{1 dist}         &  80.28 & 56.20 & 57.38 & 53.59 \\ 
\textbf{6 dist}         &  68.86 & 52.01 & 56.49 & 52.68 \\ 
\textbf{15 dist}        &  64.15 & 50.79 & 50.76 & 50.77 \\ 
\textbf{26 dist}        &  62.74 & 55.81 & 50.78 & 50.75 \\ 
\end{tabular}
\end{center}
\vspace{-4mm}
\end{table}

Another explanation for poor performance on the task is that \textit{Pathtracker} videos consist of a single channel, containing dots and markers. It is possible that model performance could be improved by engineering the dataset to disentangle dots and markers at the input, ensuring that these resources do not interfere with each other when they are encoded by models~\cite{channelgating, d2l}. To test this hypothesis, we generated new versions of \textit{Pathtracker} with three-channel videos, in which the first channel contained the start marker, the second channel contained the dots, and the third channel contained the finish marker (Fig. \ref{fig:eng}). However, I3D performed similarly on these datasets as it did on the original ones. 

We also tested whether training on optic flow encodings of \textit{Pathtracker} videos would improve I3D performance. However, this led to similar performance as when I3D was trained on the original dataset (see section \ref{sec:opticflow} in supplementary materials for details and results). 


\section{Discussion and Conclusion}
We found that I3D struggles to learn most of the \textit{Pathtracker} challenge. In particular, I3D performance was strongly affected by dot clutter, speed, and path length. Importantly, I3D \textit{is} capable of solving a baseline version of the challenge with a single distractor in relatively short (32-frame) videos. Thus, \textit{Pathtracker} stands as a novel challenge for spatiotemporal neural networks.

What mechanisms do biological visual systems rely on for appearance-free tracking? Evidence from cognitive science suggests that human observers explicitly ``trace'' the trajectories of individual dots \cite{pylyshyn2000, cavanagh2001sprites}. Similar types of tracing routines have been associated with the computations of neural circuits, and specifically the ability to transitively group features according to gestalt rules like ``good continuity'' \cite{linsley2018,roelfsema2011, raudies2011model}. We suspect that similar circuits might help on \textit{Pathtracker}, although they may need additional mechanisms for learning to detect spatiotemporal features.



Even though appearance-free tracking hasn't received much attention in computer vision to date, solving this problem holds immense promise in resolving the brittle generalization of neural networks. Indeed, a network that can solve \textit{Pathtracker} might also be expected to be able to generalize between standard object tracking benchmarks, and synthetic or real-world tracking applications, with little or no additional retraining. Furthermore, combining a good appearance-free tracker with typical apperance-based trackers might improve better performance when asked to track over long periods of time or in the presence of clutter or occlusion.

We have proposed a novel synthetic and feature agnostic tracking task inspired by cognitive psychology that challenges the long-range spatiotemporal tracking abilities of deep neural networks. Our task highlights a key limitation of the current state-of-the-art models in tracking objects without appearance cues.


\section*{Acknowledgements}
This work was supported in part by ONR (N00014-19-1-2029), NSF (IIS-1912280), NIH/NINDS (R21 NS 112743), and the ANR-3IA Artificial and Natural Intelligence Toulouse Institute (ANR-19-PI3A-0004).

Additional support provided by the Carney Institute for Brain Science, the Center for Vision Research (CVR) and the Center for Computation and Visualization (CCV) at Brown University. We acknowledge the Cloud TPU hardware resources that Google made available via the TensorFlow Research Cloud (TFRC) program as well as computing hardware supported by NIH Office of the Director grant S10OD025181.

{\small 
\bibliographystyle{ieee_fullname}
\bibliography{cvpr}
}
\clearpage
\setcounter{figure}{0}
\setcounter{table}{0}
\setcounter{section}{0}

\makeatletter 
\renewcommand{\thesection}{S\@arabic\c@section}
\renewcommand{\thefigure}{S\@arabic\c@figure}
\renewcommand{\thetable}{S\@arabic\c@table}
\makeatother

\twocolumn[  
    \begin{@twocolumnfalse}
        \begin{center}

             \Large \textbf{The Challenge of Appearance-Free Object Tracking \\ with Feedforward Neural Networks \\-- Supplementary Information --}

         \end{center}
     \end{@twocolumnfalse}
]

\section{Hyperparameter Search}
\label{sec:rgbhypersearch}

We performed an extensive hyperparameter search with variable learning rates and weight decays on all constant speed RGB datasets with mixed channels (Fig. \ref{fig:fig1}) to check if any combination of learning rate and weight decay yields better accuracy with our engineered dataset than our earlier experiments (see tables \ref{tab:rgbhs32}, \ref{tab:rgbhs64}, \ref{tab:rgbhs128}). The default learning rate and weight decay for our earlier experiments were set at $3\times10^{-4}$ and $4\times10^{-5}$ respectively (first cell in all tables). We found that barring 128-frame 26 distractor dataset, there was no significant improvement in accuracy compared to our default experiments.

\begin{table}[ht]
\caption{Evaluation accuracies from hyperparameter Search on I3D with RGB datasets of dimensions $32\times32\times32$ using given learning rates and weight decays}\label{tab:rgbhs32}
\vspace{-4mm}
\begin{center}\small

\begin{tabular}{p{0.15\linewidth}llll}
\textbf{$1$ dist.} & \multicolumn{4}{l}{\textbf{Learning Rate}} \\
\textbf{Weight Decay} & \textbf{$3\times 10^{-4}$} & \textbf{$1\times 10^{-4}$} & \textbf{$1\times 10^{-5}$} & \textbf{$1\times 10^{-6}$} \\
\textbf{$4\times 10^{-5}$} & 93.52 & 92.90 & 89.44 & 82.78 \\ 
\textbf{$1\times 10^{-6}$} & 93.88 & 93.38 & 89.60 & 82.58 \\ 
\end{tabular}

\begin{tabular}{p{0.15\linewidth}llll}
\hline
\textbf{$6$ dist.} & \multicolumn{4}{l}{\textbf{Learning Rate}} \\
\textbf{Weight Decay} & \textbf{$3\times 10^{-4}$} & \textbf{$1\times 10^{-4}$} & \textbf{$1\times 10^{-5}$} & \textbf{$1\times 10^{-6}$} \\
\textbf{$4\times 10^{-5}$} & 86.90 & 86.19 & 79.11 & 76.09 \\ 
\textbf{$1\times 10^{-6}$} & 86.30 & 86.16 & 79.90 & 77.41 \\ 
\end{tabular}

\begin{tabular}{p{0.15\linewidth}llll}
\hline
\textbf{$15$ dist.} & \multicolumn{4}{l}{\textbf{Learning Rate}} \\
\textbf{Weight Decay} & \textbf{$3\times 10^{-4}$} & \textbf{$1\times 10^{-4}$} & \textbf{$1\times 10^{-5}$} & \textbf{$1\times 10^{-6}$} \\
\textbf{$4\times 10^{-5}$} & 83.44 & 83.22 & 78.22 & 75.36 \\ 
\textbf{$1\times 10^{-6}$} & 83.59 & 82.32 & 77.79 & 75.15 \\ 
\end{tabular}

\begin{tabular}{p{0.15\linewidth}llll}
\hline
\textbf{$26$ dist.} & \multicolumn{4}{l}{\textbf{Learning Rate}} \\
\textbf{Weight Decay} & \textbf{$3\times 10^{-4}$} & \textbf{$1\times 10^{-4}$} & \textbf{$1\times 10^{-5}$} & \textbf{$1\times 10^{-6}$} \\
\textbf{$4\times 10^{-5}$} & 80.09 & 79.29 & 77.41 & 67.83 \\ 
\textbf{$1\times 10^{-6}$} & 81.03 & 79.11 & 77.50 & 70.61 \\ 
\end{tabular}

\end{center}
\end{table}

\begin{table}[ht]
\caption{Evaluation accuracies from hyperparameter Search on I3D with RGB datasets of dimensions $64\times32\times32$ using given learning rates and weight decays}\label{tab:rgbhs64}
\vspace{-4mm}
\begin{center}\small
\begin{tabular}{p{0.15\linewidth}llll}
\textbf{$1$ dist.} & \multicolumn{4}{l}{\textbf{Learning Rate}} \\
\textbf{Weight Decay} & \textbf{$3\times 10^{-4}$} & \textbf{$1\times 10^{-4}$} & \textbf{$1\times 10^{-5}$} & \textbf{$1\times 10^{-6}$} \\
\textbf{$4\times 10^{-5}$} & 80.07 & 79.55 & 73.41 & 67.16 \\ 
\textbf{$1\times 10^{-6}$} & 81.29 & 79.60 & 74.35 & 67.47 \\ 
\end{tabular}

\begin{tabular}{p{0.15\linewidth}llll}
\hline
\textbf{$6$ dist.} & \multicolumn{4}{l}{\textbf{Learning Rate}} \\
\textbf{Weight Decay} & \textbf{$3\times 10^{-4}$} & \textbf{$1\times 10^{-4}$} & \textbf{$1\times 10^{-5}$} & \textbf{$1\times 10^{-6}$} \\
\textbf{$4\times 10^{-5}$} & 66.62 & 66.25 & 62.80 & 59.34 \\ 
\textbf{$1\times 10^{-6}$} & 67.39 & 66.40 & 62.71 & 58.99 \\ 
\end{tabular}

\begin{tabular}{p{0.15\linewidth}llll}
\hline
\textbf{$15$ dist.} & \multicolumn{4}{l}{\textbf{Learning Rate}} \\
\textbf{Weight Decay} & \textbf{$3\times 10^{-4}$} & \textbf{$1\times 10^{-4}$} & \textbf{$1\times 10^{-5}$} & \textbf{$1\times 10^{-6}$} \\
\textbf{$4\times 10^{-5}$} & 65.06 & 63.75 & 61.19 & 56.38 \\ 
\textbf{$1\times 10^{-6}$} & 65.13 & 63.87 & 61.10 & 54.90 \\ 
\end{tabular}

\begin{tabular}{p{0.15\linewidth}llll}
\hline
\textbf{$26$ dist.} & \multicolumn{4}{l}{\textbf{Learning Rate}} \\
\textbf{Weight Decay} & \textbf{$3\times 10^{-4}$} & \textbf{$1\times 10^{-4}$} & \textbf{$1\times 10^{-5}$} & \textbf{$1\times 10^{-6}$} \\
\textbf{$4\times 10^{-5}$} & 64.29 & 60.96 & 58.82 & 53.13 \\ 
\textbf{$1\times 10^{-6}$} & 63.00 & 63.73 & 57.61 & 53.36 \\ 
\end{tabular}

\end{center}
\end{table}

\begin{table}[ht]
\caption{Evaluation accuracies from hyperparameter Search on I3D with RGB datasets of dimensions $128\times32\times32$ using given learning rates and weight decays}\label{tab:rgbhs128}
\vspace{-4mm}
\begin{center}\small
\begin{tabular}{p{0.15\linewidth}llll}
\textbf{$1$ dist.} & \multicolumn{4}{l}{\textbf{Learning Rate}} \\
\textbf{Weight Decay} & \textbf{$3\times 10^{-4}$} & \textbf{$1\times 10^{-4}$} & \textbf{$1\times 10^{-5}$} & \textbf{$1\times 10^{-6}$} \\
\textbf{$4\times 10^{-5}$} & 56.66 & 57.70 & 55.12 & 53.77 \\ 
\textbf{$1\times 10^{-6}$} & 58.20 & 57.80 & 55.04 & 54.27 \\ 
\end{tabular}

\begin{tabular}{p{0.15\linewidth}llll}
\hline
\textbf{$6$ dist.} & \multicolumn{4}{l}{\textbf{Learning Rate}} \\
\textbf{Weight Decay} & \textbf{$3\times 10^{-4}$} & \textbf{$1\times 10^{-4}$} & \textbf{$1\times 10^{-5}$} & \textbf{$1\times 10^{-6}$} \\
\textbf{$4\times 10^{-5}$} & 56.40 & 55.22 & 53.74 & 53.51 \\ 
\textbf{$1\times 10^{-6}$} & 56.32 & 55.73 & 53.72 & 53.34 \\ 
\end{tabular}

\begin{tabular}{p{0.15\linewidth}llll}
\hline
\textbf{$15$ dist.} & \multicolumn{4}{l}{\textbf{Learning Rate}} \\
\textbf{Weight Decay} & \textbf{$3\times 10^{-4}$} & \textbf{$1\times 10^{-4}$} & \textbf{$1\times 10^{-5}$} & \textbf{$1\times 10^{-6}$} \\
\textbf{$4\times 10^{-5}$} & 56.54 & 55.47 & 52.82 & 52.63 \\ 
\textbf{$1\times 10^{-6}$} & 56.55 & 55.86 & 53.51 & 52.59 \\ 
\end{tabular}

\begin{tabular}{p{0.15\linewidth}llll}
\hline
\textbf{$26$ dist.} & \multicolumn{4}{l}{\textbf{Learning Rate}} \\
\textbf{Weight Decay} & \textbf{$3\times 10^{-4}$} & \textbf{$1\times 10^{-4}$} & \textbf{$1\times 10^{-5}$} & \textbf{$1\times 10^{-6}$} \\
\textbf{$4\times 10^{-5}$} & 50.81 & 55.11 & 51.91 & 51.35 \\ 
\textbf{$1\times 10^{-6}$} & 53.65 & 56.06 & 52.49 & 51.67 \\ 
\end{tabular}

\end{center}
\end{table}

\newpage
\section{Experiments with Optic Flow}
\label{sec:opticflow}

\begin{figure*}[ht]
\begin{center}\small
\includegraphics[width=.67\textwidth]{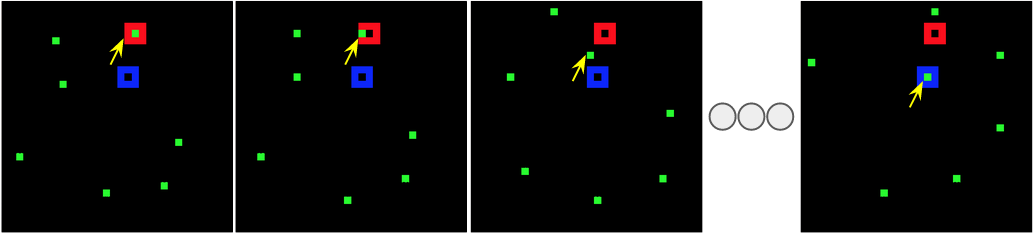}
\end{center}
\vspace{-4mm}
\caption{Positive example from the engineered dataset of \textit{Pathtracker} challenge. The markers and dots are spread across multiple channels. The starting marker is in the first channel, the dots are in second channel, while the ending marker is in the last channel. Everything else is same as our original dataset in Figure~\ref{fig:fig1}. See section \ref{sec:challenge} for description on challenge.}
\vspace{-4mm}
\label{fig:eng}
\end{figure*}

As done in \cite{carreira2017}, we calculated optic flow for all the datasets using TV-L1 algorithm \cite{tvl1, tvl1zach}. We engineered our dataset (see Figure~\ref{fig:eng}) to check if spreading features across different channels could help improve the performance of I3D \cite{channelgating, d2l}. To this end, we put the starting marker in the first channel, dots in the second, and ending marker in the last channel. We further used OpenCV's TV-L1 implementation \cite{tvl1zach} to calculate optic flow on our videos. We used two channels from the output given by the TV-L1 algorithm, and appended one channel from the raw dataset to simulate the two-stream network described in \cite{carreira2017} in their RGB+Flow experiments. We found that engineered dataset combined with optic flow was unable to rescue the performance of I3D on our Pathtracker challenge (see table \ref{tab:of1}). In a few cases, we even observed a drop in accuracy compared to our earlier experiments without optic flow (compare with table \ref{tab:table1}). 

\begin{table}[ht]
\caption{Evaluation accuracy of I3D with optic flow with variation in the number of frames.}
\vspace{-4mm}
\label{tab:of1}
\begin{center}\small
\begin{tabular}{llll}
&  \multicolumn{3}{l}{\textbf{Video dimension (D,H,W)}} \\
\textbf{Distractors}               & \textbf{32,32,32}                & \textbf{64,32,32}                       & \textbf{128,32,32}                       \\
\textbf{1 dist}         & 93.35              & 78.77                      & 51.20                      \\ 
\textbf{6 dist}         & 83.71              & 65.94                     & 51.84                      \\ 
\textbf{15 dist}        & 79.59              & 62.21                     & 51.03                      \\ 
\textbf{26 dist}        & 78.50              & 51.55 & 50.85 \\ 
\end{tabular}
\end{center}
\vspace{-4mm}
\end{table}

\newpage
\section{Hyperparameter Search with Optic Flow}
\label{sec:hs}
We performed an extensive hyperparameter search with variable learning rates and weight decays on all constant speed datasets with separate channels and optic flow to check if any combination of learning rate and weight decay yields better accuracy with our engineered dataset than our earlier experiments (see tables \ref{tab:hs32}, \ref{tab:hs64}, \ref{tab:hs128}). The default learning rate and weight decay for our earlier experiments were set at $3\times10^{-4}$ and $4\times10^{-5}$ respectively (first cell in all tables). We found that barring 64-frame 26 distractor dataset, there was no significant improvement in accuracy compared to our default experiments. Different combinations of hyperparameters coupled with optic flow were again insufficient in rescuing performance of I3D on our Pathtracker challenge.

\begin{table}[ht]
\caption{Evaluation accuracies from hyperparameter Search on I3D with optic flow datasets of dimensions $32\times32\times32$ using given learning rates and weight decays}
\vspace{-4mm}
\label{tab:hs32}
\begin{center}\small

\begin{tabular}{p{0.15\linewidth}llll}
\textbf{$1$ dist.} & \multicolumn{4}{l}{\textbf{Learning Rate}} \\
\textbf{Weight Decay} & \textbf{$3\times 10^{-4}$} & \textbf{$1\times 10^{-4}$} & \textbf{$1\times 10^{-5}$} & \textbf{$1\times 10^{-6}$} \\
\textbf{$4\times 10^{-5}$} & 93.35 & 93.00 & 89.55 & 78.73 \\ 
\textbf{$1\times 10^{-6}$} & 93.50 & 93.50 & 88.63 & 80.17 \\ 
\end{tabular}

\begin{tabular}{p{0.15\linewidth}llll}
\hline
\textbf{$6$ dist.} & \multicolumn{4}{l}{\textbf{Learning Rate}} \\
\textbf{Weight Decay} & \textbf{$3\times 10^{-4}$} & \textbf{$1\times 10^{-4}$} & \textbf{$1\times 10^{-5}$} & \textbf{$1\times 10^{-6}$} \\
\textbf{$4\times 10^{-5}$} & 83.71 & 84.72 & 78.61 & 72.58 \\ 
\textbf{$1\times 10^{-6}$} & 83.98 & 84.08 & 78.03 & 72.64 \\ 
\end{tabular}

\begin{tabular}{p{0.15\linewidth}llll}
\hline
\textbf{$15$ dist.} & \multicolumn{4}{l}{\textbf{Learning Rate}} \\
\textbf{Weight Decay} & \textbf{$3\times 10^{-4}$} & \textbf{$1\times 10^{-4}$} & \textbf{$1\times 10^{-5}$} & \textbf{$1\times 10^{-6}$} \\
\textbf{$4\times 10^{-5}$} & 79.59 & 78.39 & 76.13 & 68.39 \\ 
\textbf{$1\times 10^{-6}$} & 79.63 & 78.86 & 76.00 & 65.27 \\ 
\end{tabular}

\begin{tabular}{p{0.15\linewidth}llll}
\hline
\textbf{$26$ dist.} & \multicolumn{4}{l}{\textbf{Learning Rate}} \\
\textbf{Weight Decay} & \textbf{$3\times 10^{-4}$} & \textbf{$1\times 10^{-4}$} & \textbf{$1\times 10^{-5}$} & \textbf{$1\times 10^{-6}$} \\
\textbf{$4\times 10^{-5}$} & 78.50 & 78.42 & 74.44 & 62.98 \\ 
\textbf{$1\times 10^{-6}$} & 78.54 & 78.40 & 73.53 & 59.92 \\ 
\end{tabular}

\end{center}
\vspace{-4mm}
\end{table}

\begin{table}[ht]
\caption{Evaluation accuracies from hyperparameter Search on I3D with optic flow datasets of dimensions $64\times32\times32$ using given learning rates and weight decays}
\vspace{-4mm}
\label{tab:hs64}
\begin{center}\small

\begin{tabular}{p{0.15\linewidth}llll}
\textbf{$1$ dist.} & \multicolumn{4}{l}{\textbf{Learning Rate}} \\
\textbf{Weight Decay} & \textbf{$3\times 10^{-4}$} & \textbf{$1\times 10^{-4}$} & \textbf{$1\times 10^{-5}$} & \textbf{$1\times 10^{-6}$} \\
\textbf{$4\times 10^{-5}$} & 78.77 & 77.81 & 70.27 & 62.20 \\ 
\textbf{$1\times 10^{-6}$} & 79.14 & 78.02 & 70.19 & 61.45 \\ 
\end{tabular}

\begin{tabular}{p{0.15\linewidth}llll}
\hline
\textbf{$6$ dist.} & \multicolumn{4}{l}{\textbf{Learning Rate}} \\
\textbf{Weight Decay} & \textbf{$3\times 10^{-4}$} & \textbf{$1\times 10^{-4}$} & \textbf{$1\times 10^{-5}$} & \textbf{$1\times 10^{-6}$} \\
\textbf{$4\times 10^{-5}$} & 65.94 & 65.93 & 59.56 & 55.78 \\ 
\textbf{$1\times 10^{-6}$} & 66.58 & 66.06 & 60.64 & 56.91 \\ 
\end{tabular}

\begin{tabular}{p{0.15\linewidth}llll}
\hline
\textbf{$15$ dist.} & \multicolumn{4}{l}{\textbf{Learning Rate}} \\
\textbf{Weight Decay} & \textbf{$3\times 10^{-4}$} & \textbf{$1\times 10^{-4}$} & \textbf{$1\times 10^{-5}$} & \textbf{$1\times 10^{-6}$} \\
\textbf{$4\times 10^{-5}$} & 62.21 & 61.82 & 58.05 & 54.63 \\ 
\textbf{$1\times 10^{-6}$} & 62.20 & 62.39 & 58.04 & 53.66 \\ 
\end{tabular}

\begin{tabular}{p{0.15\linewidth}llll}
\hline
\textbf{$26$ dist.} & \multicolumn{4}{l}{\textbf{Learning Rate}} \\
\textbf{Weight Decay} & \textbf{$3\times 10^{-4}$} & \textbf{$1\times 10^{-4}$} & \textbf{$1\times 10^{-5}$} & \textbf{$1\times 10^{-6}$} \\
\textbf{$4\times 10^{-5}$} & 51.55 & 61.75 & 54.62 & 52.59 \\ 
\textbf{$1\times 10^{-6}$} & 62.07 & 60.20 & 56.26 & 52.36 \\ 
\end{tabular}

\end{center}
\vspace{-4mm}
\end{table}

\begin{table}[ht]
\caption{Evaluation accuracies from hyperparameter Search on I3D with optic flow datasets of dimensions $128\times32\times32$ using given learning rates and weight decays}
\vspace{-4mm}
\label{tab:hs128}
\begin{center}\small

\begin{tabular}{p{0.15\linewidth}llll}
\textbf{$1$ dist.} & \multicolumn{4}{l}{\textbf{Learning Rate}} \\
\textbf{Weight Decay} & \textbf{$3\times 10^{-4}$} & \textbf{$1\times 10^{-4}$} & \textbf{$1\times 10^{-5}$} & \textbf{$1\times 10^{-6}$} \\
\textbf{$4\times 10^{-5}$} & 51.20 & 57.29 & 54.56 & 53.98 \\ 
\textbf{$1\times 10^{-6}$} & 54.66 & 57.22 & 55.10 & 53.60 \\ 
\end{tabular}

\begin{tabular}{p{0.15\linewidth}llll}
\hline
\textbf{$6$ dist.} & \multicolumn{4}{l}{\textbf{Learning Rate}} \\
\textbf{Weight Decay} & \textbf{$3\times 10^{-4}$} & \textbf{$1\times 10^{-4}$} & \textbf{$1\times 10^{-5}$} & \textbf{$1\times 10^{-6}$} \\
\textbf{$4\times 10^{-5}$} & 51.84 & 54.34 & 53.45 & 52.24 \\ 
\textbf{$1\times 10^{-6}$} & 56.30 & 54.26 & 52.69 & 52.11 \\ 
\end{tabular}

\begin{tabular}{p{0.15\linewidth}llll}
\hline
\textbf{$15$ dist.} & \multicolumn{4}{l}{\textbf{Learning Rate}} \\
\textbf{Weight Decay} & \textbf{$3\times 10^{-4}$} & \textbf{$1\times 10^{-4}$} & \textbf{$1\times 10^{-5}$} & \textbf{$1\times 10^{-6}$} \\
\textbf{$4\times 10^{-5}$} & 51.03 & 54.49 & 52.74 & 52.07 \\ 
\textbf{$1\times 10^{-6}$} & 52.96 & 55.43 & 53.03 & 51.82 \\ 
\end{tabular}

\begin{tabular}{p{0.15\linewidth}llll}
\hline
\textbf{$26$ dist.} & \multicolumn{4}{l}{\textbf{Learning Rate}} \\
\textbf{Weight Decay} & \textbf{$3\times 10^{-4}$} & \textbf{$1\times 10^{-4}$} & \textbf{$1\times 10^{-5}$} & \textbf{$1\times 10^{-6}$} \\
\textbf{$4\times 10^{-5}$} & 50.85 & 50.81 & 51.80 & 50.96 \\ 
\textbf{$1\times 10^{-6}$} & 50.77 & 53.61 & 51.66 & 51.63 \\ 
\end{tabular}

\end{center}
\vspace{-4mm}
\end{table}

\end{document}